\documentclass{article}

\usepackage{microtype}
\usepackage{graphicx}
\usepackage{subcaption}
\usepackage{booktabs} % for professional tables

\usepackage{hyperref}

\usepackage[preprint]{icml2026}

\usepackage{amsmath}
\usepackage{amssymb}
\usepackage{mathtools}
\usepackage{amsthm}
\usepackage{graphicx}
\usepackage{booktabs}
\usepackage{tabularx}
\usepackage{multirow}
\usepackage{array}
\usepackage{makecell}
\usepackage[table]{xcolor}
\usepackage{pifont}
\usepackage{bm}
\usepackage{xurl}

\usepackage{xspace}
\newcommand{\ARCH}{HypeNet\xspace}
\newcommand{\DISTILL}{HALO\xspace}

\newcommand{\citepnoyear}[1]{(\citeauthor{#1})}

\usepackage[capitalize,noabbrev]{cleveref}

\theoremstyle{plain}

\theoremstyle{definition}

\theoremstyle{remark}

\usepackage[textsize=tiny]{todonotes}

\icmltitlerunning{Hybrid Linear Attention Done Right}

\begin{document}

\twocolumn[
  % \icmltitle{\ARCH: Fast Attention Hybridization for Extremely Long Contexts \\
    % International Conference on Machine Learning (ICML 2026)}
  \icmltitle{Hybrid Linear Attention Done Right:\\Efficient Distillation and Effective Architectures for Extremely Long Contexts}

  % It is OKAY to include author information, even for blind submissions: the
  % style file will automatically remove it for you unless you've provided
  % the [accepted] option to the icml2026 package.

  % List of affiliations: The first argument should be a (short) identifier you
  % will use later to specify author affiliations Academic affiliations
  % should list Department, University, City, Region, Country Industry
  % affiliations should list Company, City, Region, Country

  % You can specify symbols, otherwise they are numbered in order. Ideally, you
  % should not use this facility. Affiliations will be numbered in order of
  % appearance and this is the preferred way.
  \icmlsetsymbol{equal}{*}

  \begin{icmlauthorlist}
    \icmlauthor{Yingfa Chen}{equal,thu}
    \icmlauthor{Zhen Leng Thai}{equal,thu}
    \icmlauthor{Zihan Zhou}{modelbest}
    \icmlauthor{Zhu Zhang}{modelbest}
    \icmlauthor{Xingyu Shen}{thu}\\
    \icmlauthor{Shuo Wang}{thu}
    \icmlauthor{Chaojun Xiao}{thu}
    %\icmlauthor{}{sch}
    \icmlauthor{Xu Han}{thu}
    \icmlauthor{Zhiyuan Liu}{thu}
    %\icmlauthor{}{sch}
    %\icmlauthor{}{sch}
  \end{icmlauthorlist}

  \icmlaffiliation{thu}{NLP Group, DCST, IAI, Tsinghua University}
  \icmlaffiliation{modelbest}{ModelBest Inc., Beijing, China}
  % \icmlaffiliation{sch}{Tsinghua University, IAI, Beijing, China}

  % \icmlcorrespondingauthor{Yingfa Chen}{xxx@xxx.com}
  % \icmlcorrespondingauthor{Zhen Leng Thai}{xxx@xxx.com}

    \begin{center}
        \textsuperscript{1}NLP Group, DCST, IAI, BNRIST, Tsinghua University, Beijing, China \quad 
        \textsuperscript{2}OpenBMB\\
        {\tt \{chenyingfa1999, thaizhenleng123\}@gmail.com }\\
        {\tt wangshuo.thu@gmail.com, han-xu@tsinghua.edu.cn }
    \end{center}

  % You may provide any keywords that you find helpful for describing your
  % paper; these are used to populate the "keywords" metadata in the PDF but
  % will not be shown in the document
  \icmlkeywords{Machine Learning, ICML}

  \vskip 0.3in
]

% this must go after the closing bracket ] following \twocolumn[ ...

% This command actually creates the footnote in the first column listing the
% affiliations and the copyright notice. The command takes one argument, which
% is text to display at the start of the footnote. The \icmlEqualContribution
% command is standard text for equal contribution. Remove it (just {}) if you
% do not need this facility.

% Use ONE of the following lines. DO NOT remove the command.
% If you have no special notice, KEEP empty braces:
% \printAffiliationsAndNotice{}  % no special notice (required even if empty)
% Or, if applicable, use the standard equal contribution text:
% \printAffiliationsAndNotice{\icmlEqualContribution}

\begin{abstract}

Hybrid Transformer architectures, which combine softmax attention blocks and recurrent neural networks (RNNs), have shown a desirable performance-throughput tradeoff for long-context modeling, but their adoption and studies are hindered by the prohibitive cost of large-scale pre-training from scratch.
Some recent studies have shown that pre-trained softmax attention blocks can be converted into RNN blocks through parameter transfer and knowledge distillation.
However, these transfer methods require substantial amounts of training data (more than 10B tokens), and the resulting hybrid models also exhibit poor long-context performance, which is the scenario where hybrid models enjoy significant inference speedups over Transformer-based models.
In this paper, we present \DISTILL (Hybrid Attention via Layer Optimization), a pipeline for distilling Transformer models into RNN-attention hybrid models. We then present \ARCH, a hybrid architecture with superior length generalization enabled by a novel position encoding scheme (named HyPE) and various architectural modifications.
We convert the Qwen3 series into \ARCH using \DISTILL, achieving performance comparable to the original Transformer models while enjoying superior long-context performance and efficiency. The conversion requires just 2.3B tokens, less than 0.01\% of their pre-training data\footnote{The code and model checkpoints can be found at: \url{https://github.com/THUNLP/hybrid-linear-attention}.}.

\end{abstract}

\begingroup
\renewcommand{\thefootnote}{\fnsymbol{footnote}}
\footnotetext[1]{Equal contributions.}
% \footnotetext[2]{Corresponding authors.}
\endgroup

\section{Introduction}

Transformer-based language models~\citep{transformer} rely on softmax attention blocks, which have a quadratic complexity with respect to the context length, making them prohibitively expensive for long contexts. In contrast, recurrent neural networks (RNNs) such as linear attention~\citep{transformers-are-rnns} and state space models~\citep{mamba} are much faster for long-context modeling due to their linear complexity.
However, pure RNN models with fixed-size states generally underperform softmax attention, particularly on recall-intensive tasks~\citep{repeat-after-me, gdn}. To address this gap, there is a surge in interest in hybrid architectures that interleave attention and RNN layers\footnote{We hereby use \textit{hybrid architectures/models} to refer to architectures/models that consist of softmax attention and RNN layers.}, achieving a favorable tradeoff between model performance and inference throughput~\citep{jamba, minimax-01, qwen3-next, kimi-linear, nemotron-nano-3}.

\begin{figure*}[!t]
    \centering
    \includegraphics[width=\linewidth]{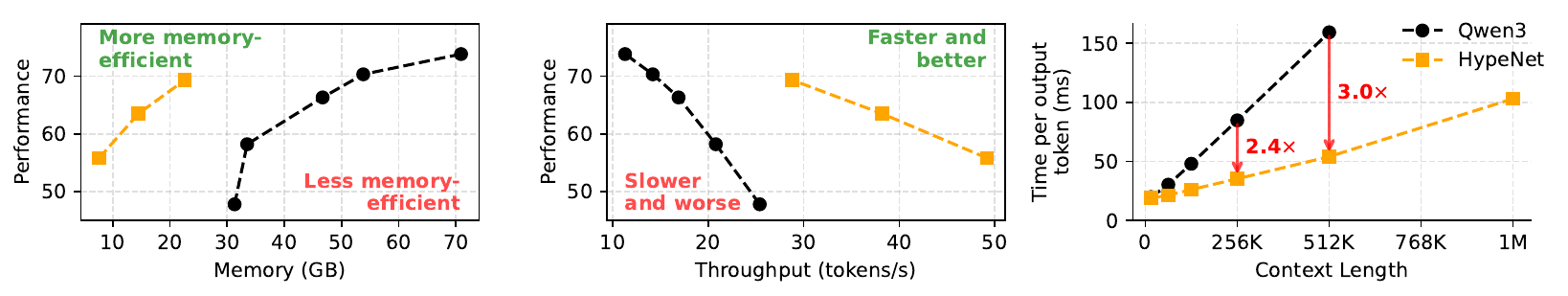}
    \caption{Left \& center: the performance-efficiency tradeoff of our model, \textbf{\textcolor{orange}{\ARCH}}, versus the \textbf{Qwen3} series, measured with 128K context length and BFloat16 precision. Right: the time per output token of the 1.7B models across different context lengths. For 1M context length, the Qwen3 model runs out of GPU memory. \ARCH is converted from Qwen3 using our distillation procedure, \DISTILL, and has better performance-efficiency tradeoff than Qwen3.}
    \label{fig:performance-vs-cost}
\end{figure*}

Hybrid architectures are typically pre-trained from scratch at a large scale~\cite{qwen3-next,nemotron-nano-3}, placing them beyond the reach of most academic research teams. 
Hence, some works focus on distilling pre-trained Transformer models into hybrid architectures~\citep{jet-nemotron, RAD, mamba-in-the-llama}. These distillation methods use far fewer training tokens and produce hybrid models that are comparable to their Transformer counterparts on various common-sense reasoning (CSR) tasks. Although distilled hybrid models typically underperform those trained from scratch, they are valuable since they allow teams without resources to scale up pre-training to validate research ideas.

However, these distillation methods still suffer from two critical limitations. (1)~Most distillation methods still require tens to hundreds of billions of training tokens, which is still out of reach for most teams in academia. (2)~While the resulting hybrid models have short-context performance comparable to Transformer models, they exhibit severe performance degradation on long-context tasks, which is precisely the scenario where they are preferred over Transformer models.

To address these challenges, we first propose \DISTILL (Hybrid Attention via Layer Optimization), a novel cross-architecture distillation procedure for converting pre-trained Transformer models into hybrid models. Notably, \DISTILL~involves an efficient \textit{attention layer selection} method for determining which attention layers to keep unconverted to ensure the best long-context performance. Then, we propose Hybrid Position Encoding (HyPE), a position encoding scheme with strong length generalization, specifically designed for hybrid architectures. In addition to HyPE, we propose a series of architectural improvements, validated with careful ablation experiments on models with over 1B parameters. The combination of these improvements results in \ARCH, a series of hybrid models converted from the Qwen3 series, with a much better performance-throughput tradeoff, as shown in Figure~\ref{fig:performance-vs-cost}.

Our contributions can be summarized as follows:
\begin{itemize}
    \item We develop a novel cross-architecture distillation procedure that converts Transformer models into attention-RNN hybrid models using fewer than 3B tokens, thereby significantly improving the model's efficiency in long-context scenarios.
    \item We present HyPE, a novel position-encoding scheme that combines RoPE~\citep{rope} and NoPE~\citep{nope}, designed for hybrid models. Coupled with an attention scaling mechanism, HyPE achieves superior length generalization.
    \item Based on HyPE, we propose \ARCH, a novel hybrid architecture that incorporates multiple architectural improvements when converting from a pre-trained Transformer model.
\end{itemize}

\section{Related Works}
\label{sec:related-works}

\begin{table}[!t]
    \centering
    \footnotesize
    \caption{Existing attention-to-hybrid distillation methods and their release date and training tokens required.}
    \begin{tabular}{lcr}
        \toprule
        \textbf{Method} & \textbf{Date} & \textbf{Tokens} \\
        \midrule
        Mamba-in-the-Llama \citepnoyear{mamba-in-the-llama} 
            & Aug. 2024 &  20B \\
        SMART \citepnoyear{zebrallama}            
            & May 2025  & $>$7B\\
        RAD \citepnoyear{RAD}                     
            & May 2025  & 20B \\
        Jet-Nemotron \citepnoyear{jet-nemotron}   
            & Aug. 2025 & 400B  \\
        KL-LS \citepnoyear{kl-guided-layer-selection} 
            & Dec. 2025 & 25B \\
        \midrule
        \DISTILL (ours) 
            & Jan. 2026 & 2.3B \\
        \bottomrule
    \end{tabular}
    \label{tab:previous-works}
\end{table}

\paragraph{RNN-Attention Hybrid Models}

State-of-the-art hybrid models with up to hundreds of billions of parameters have exhibited performance comparable to standard Transformers on both commonsense reasoning and recall-intensive tasks~(e.g., needle-in-a-haystack~(NIAH)~\citep{ruler}) while being more efficient for processing long contexts~\citep{jamba, minimax-01, qwen3-next, kimi-linear, nemotron-nano-3}. Despite their impressive performance, there are rather few publicly available hybrid models with frontier-level performance, because pre-training from scratch is prohibitively expensive for most teams. To avoid this training cost, we focus on distilling pre-trained Transformer models into hybrid models.

\paragraph{Position Encoding in Hybrid Models}

Current, RoPE~\citep{rope} has become the \textit{de facto} standard position encoding~(PE) for Transformer models~\citep{qwen3, llama3}. On the other hand, RNNs usually encode positional information through decay/transition matrices, and do not employ RoPE~\citep{mamba2,gdn}. This has remained the case for hybrid models, which means attention layers adopt RoPE while RNN layers do not (i.e., RNNs use NoPE)~\citep{qwen3-next,minimax-01}. Recently, SWAN-GPT~\citep{swan-gpt} has shown promising long-context generalization by combining RoPE in sliding window attention layers and NoPE in full attention layers, but it is not a hybrid model. Concurrent to this paper, Kimi-Linear~\citep{kimi-linear} has adopted NoPE in both attention and RNN layers. In contrast, our model employs a novel PE scheme and achieves better long-context performance than typical PE methods found in existing hybrid models.

\paragraph{Distilling Transformers into Hybrid Models} 

Many works focus on converting Transformers into pure RNN models via distillation~\citep{T2R,mohawk,lolcats,radlads}, but converting Transformers into hybrid models remains underexplored. When distilling into hybrids, choosing which attention layer to convert to RNN is critical for maintaining performance, especially for tasks that are hard to handle with RNN layers. \citet{mamba-in-the-llama} adopt a simple pipeline and attention layer selection scheme and show severe performance degradation. More recent works choose which attention layers to retain more sophistically. \citet{zebrallama} use the output distribution shift when replacing an attention layer with an RNN layer to determine the importance of attention layers. \citet{RAD} propose a redundancy metric for determining importance, and \citet{jet-nemotron} rely on the performance drop on certain tasks. Finally, KL-guided layer selection~(KL-LS)~\citep{kl-guided-layer-selection}, a concurrent work, proposes using KL-divergence from the teacher model as the importance metric and requires a thorough search that repeatedly reruns a distillation process for every layer. 

Table~\ref{tab:previous-works} lists previous distillation works. These works typically use more than 10B training tokens and have poor recall performance compared to Transformer models, especially on long contexts. In contrast, our distillation procedure requires just 2.3B tokens, and our architecture has much stronger long-context performance thanks to its superior length generalization.

\begin{figure*}[!t]
    \centering
    \includegraphics[width=0.9\linewidth]{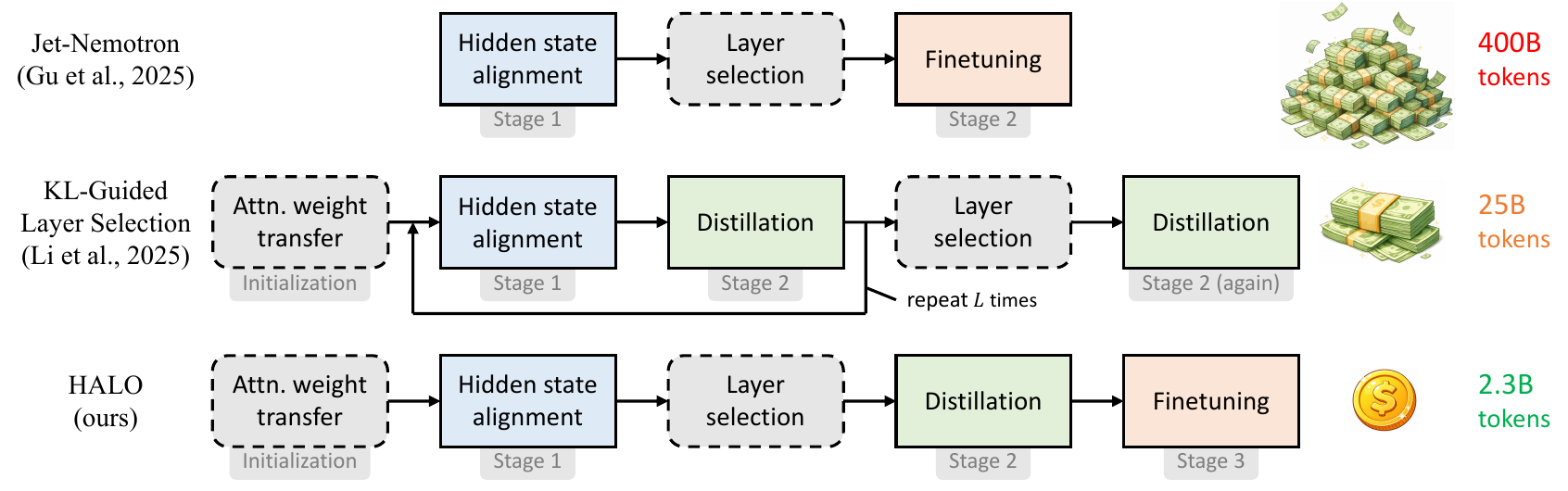}
    \caption{Various pipelines for converting Transformer models into hybrid models. The boxes with dotted lines represent training-free stages, while those with solid lines represent training stages. \DISTILL is much more data-efficient than prior methods.}
    \label{fig:pipeline}
\end{figure*}

\section{Preliminaries}
\label{sec:preliminaries}

\paragraph{Notations}

All models involved in this study, including both Transformer and hybrid models, consist of a stack of $L$ layers, and the $l$-th layer can be formalized as
\begin{equation}
    \begin{aligned}
        \mathbf H^{(l)} &= \text{Mixer}^{(l)} \left(\mathbf X^{(l-1)} \right) + \mathbf X^{(l-1)},\\
        \mathbf X^{(l)} &= \text{MLP} ^{(l)} \left( \mathbf H^{(l)} \right) + \mathbf H ^{(l)},
    \end{aligned}
\end{equation}
where $\mathbf X^{(l)} = \begin{bmatrix} \mathbf x_1^\top, \cdots, \mathbf x_T^\top \end{bmatrix}^\top \in \mathbb R^{T\times d}$ denotes the $T$ $d$-dimensional output embeddings. 
In an RNN-attention hybrid model, the set of attention layers is specified by $\mathcal I_\text{attn}\in \{l_{\text{attn},i} \mid i=1,\cdots,L_\text{attn}\}$, where $L_\text{attn}$ is the number of attention layers and $l_{\text{attn}, i} \in\{1,\cdots, L\}$ is the index of the $i$-th attention layer. The mixers are defined as
\begin{equation}
    \begin{aligned}
        \text{Mixer}^{(l)} = \begin{cases}
            \text{ATTN}^{(l)} & \text{if } l \in \mathcal I_\text{attn}, \\
            \text{RNN}^{(l)} & \text{otherwise}.
        \end{cases}
    \end{aligned}
\end{equation}

\paragraph{Softmax Attention Layers}

In Transformer, the mixer layer uses softmax attention, which can be written as\footnote{Here, we ignore the multi-head mechanism for simplicity.}
\begin{equation}
    \begin{aligned}
        \mathbf Q &= \mathbf X \mathbf W_q, \quad \mathbf K = \mathbf X \mathbf W_k, \quad \mathbf V =\mathbf X \mathbf W_v,\\
        \mathbf Y &= \text{softmax}\left( \frac {1} {\sqrt{d_h}} \mathbf Q \mathbf K^\top \odot \mathbf M \right) \mathbf V \mathbf W_o^\top,
    \end{aligned}
\end{equation}
where $\mathbf W_q,\mathbf W_k, \mathbf W_v, \mathbf W_o \in \mathbb R^{d\times d_h}$ are learnable parameters, and $\mathbf M$ is the attention mask. We use row-vector representation, so $\mathbf x ^\top \mathbf x$ denotes an outer product.

\paragraph{Modern RNN Layers}

There are many variants of RNN layers, but we focus on RNNs that can be written as
\begin{align}
    \mathbf q_t &= \mathbf x_t \mathbf W_q, \quad \mathbf k_t = \mathbf x_t \mathbf W_k, \quad \mathbf v_t = \mathbf x_t \mathbf W_v, \\
    \mathbf S_{t} &= \mathbf F_{t} \mathbf S_{t-1} + \mathbf k_t^\top \mathbf v_t \in \mathbb R^{d_h\times d_h}, \label{eq:update-rule} \\
    \mathbf y_{t} &= \mathbf q_{t} \mathbf S_{t} \mathbf W_o^\top \in \mathbb R^{d}, \label{eq:query-rule}
\end{align}
where $\mathbf F_t\in\mathbb R^{d_h\times d_h}$ is named the \textit{transition matrix} and is a function of $\mathbf x_t$. The above formulas include state-of-the-art RNN variants such as Mamba2~\citep{mamba2}, Gated DeltaNet~\citep{gdn}, etc. To enable fast parallelization, $\mathbf F_t$ is typically a diagonal matrix or rank-1 matrix~\citep{gla, gdn}. $\mathbf S_t$ is named the \textit{recurrent state}\footnote{Also named \textit{hidden state} in some papers.}, and Eq.~(\ref{eq:update-rule}) and (\ref{eq:query-rule}) are named the \textit{update rule} and the \textit{query rule}, respectively.

\subsection{The Impact of Attention Layer Selection when Distilling Transformers into Hybrids}

When distilling Transformer models into hybrid models, one important question is \textit{how to select which attention layers to remain unconverted}, i.e., how to determine the optimal $\mathcal I_\text{attn}$ for maximizing model performance, without increasing the number of attention layers $\vert \mathcal I_\text{attn}\vert$ (since efficiency is negatively correlated with $\vert \mathcal I_\text{attn} \vert$). Previous works have identified that RNN models underperform attention models on recall-intensive tasks~\citep{gdn,StateX,repeat-after-me}; thus, our objective is to identify which attention layers are most important for modeling recall abilities and leave them unconverted.

\subsection{The Importance of Position Encoding for Language Modeling and Length Generalization}
\label{sec:importance-of-position-encoding}

For attention-based models, it is common to inject positional information into the model via RoPE, which applies a position-dependent rotation to $\mathbf Q$ and $\mathbf K$. Although RoPE typically improves language modeling performance of Transformer models, attention without RoPE (a.k.a., NoPE), exhibits superior training-free length generalization~\citep{nope, wang2024lengthgeneralizationcausaltransformers, swan-gpt}. Length generalization is also important for long-context post-training because models with better length generalization are more data-efficient~\cite{yarn}. 

In contrast, RNNs are inherently position-aware through the state transition $\mathbf F_t$ in their update rule. Therefore, most existing RNN models employ NoPE. However, the language modeling performance and length generalization of RNNs are sensitive to the structure and parameterization of the update rule~\citep{stuffed-mamba,gla}. Thus, in hybrid models, achieving strong performance and length generalization requires careful synergy between the update rule~(and/or PE) RNN layers and the PE in attention layers.

\section{\DISTILL: An Efficient Pipeline to Distill Transformers into Hybrids}
\label{sec:conversion}

Our conversion procedure, \DISTILL, is an adoption and improvement of RADLADS~\citep{radlads}, a distillation method that converts Transformer models into pure RNN models~\citep{rwkv7}. Figure~\ref{fig:pipeline} shows an overview of \DISTILL. It consists of an attention weight transfer process, three training stages, and an attention layer selection process. Appendix~\ref{sec:appendix-training-configs} shows the training configuration of each stage in \DISTILL.

\subsection{Initialization Stage: Attention Weight Transfer}

Given a Transformer model consisting entirely of attention layers, for each attention layer $\text{ATTN}^{(l)}(\cdot)$, we use its configuration and pre-trained projection weights $\left(\mathbf W_q, \mathbf W_k, \mathbf W_v, \mathbf W_o\right)$ to instantiate an RNN layer $\text{RNN}^{(l)}(\cdot)$. If an RNN layer has other modules that cannot be covered by the weights of the attention layer, we initialize the weights of these modules using the empirical implementation of RNN layers.

\subsection{Stage 1: Hidden State Alignment}
\label{sec:stage1}

We train each instantiated RNN layer independently by minimizing the mean squared error (MSE) between its output hidden states and the attention layer used to instantiate it:
\begin{equation}
    \begin{aligned}
        \mathcal L_\text{stage 1}^{(l)} = \text{MSE}\left(\mathbf Y^{(l)}_{\text{teacher}}, \text{RNN}^{(l)}\left(\mathbf X^{(l-1)} \right) \right),
    \end{aligned}
\end{equation}
where $\mathbf Y_\text{teacher}^{(l)}$ is the output of the $l$-th attention layer in the attention-only teacher model.
During the alignment process, only the RNN layers are trained, and all other weights are frozen.
After stage 1, each attention layer has a student RNN layer that can potentially replace it.

\begin{figure*}[t]
    \centering
    \includegraphics[width=\linewidth]{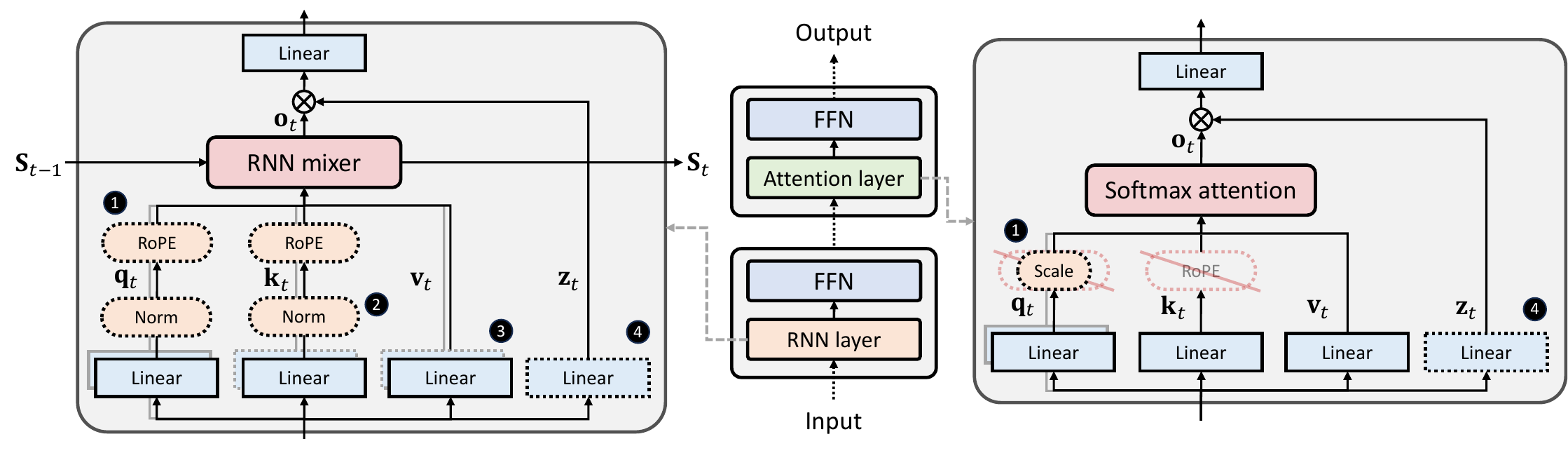}
    \caption{Illustration of \ARCH. The architectural modifications introduced during \DISTILL are marked with \ding{202}, \ding{203}, \ding{204}, and \ding{205}. \textcolor{red!80}{Red dotted lines} indicate components that are removed during \DISTILL, black dotted lines indicate components that are added.}
    \label{fig:arch}
\end{figure*}

\subsection{Attention Layer Selection}
\label{sec:attn-layer-selection}

Here, we perform attention layer selection to determine $\mathcal I_\text{attn}$.
We propose to select attention layers that, when replaced by RNN layers, exhibit \textbf{a large drop in recall performance and a small drop in CSR}. Let $M^{(i)}$ denote the original model but with the $i$-th layer replaced with the corresponding RNN layer from stage 1. Let $\mathcal R(M), \mathcal C(M)\in [0, 1]$ denote the recall and CSR performance of the model $M$, then, the importance score of each attention layer is
\begin{equation}
    \begin{aligned}
        s_i 
        = \frac{\max_i\left[  \mathcal R \left(M^{(i)} \right) \right] - \mathcal R \left(M^{(i)} \right) }
            {\max_i \left[\mathcal C\left( M^{(i)} \right) \right] - \mathcal C \left(M^{(i)} \right) + \epsilon},
    \end{aligned}
    \label{eq:layer-importance}
\end{equation}
where $\epsilon= 10^{-6}$ is a small constant to avoid division by zero. Finally, we simply pick the Top-$k$ most important attention layer as
\begin{equation}
    \begin{aligned}
        \mathcal I_\text{attn} = \underset{i}{\text{Top-}k} (s_i).
    \end{aligned}
\end{equation}
Based on \citet{wang2025systematicanalysishybridlinear}, we always use $k=\lfloor L/4 \rfloor$ in this paper, which means that 25\% of the layers in the final model are attention layers. The actual layer indices $\mathcal{I}_\text{attn}$ selected by our approach are reported in Appendix~\ref{sec:appendix-model-configs}.

\subsection{Stage 2: Knowledge Distillation}
\label{sec:stage2}

In stage 2, we construct the final hybrid model $f_\text{hybrid}$ using $\mathcal I_\text{attn}$ and conduct standard end-to-end knowledge distillation, with the original Transformer model $f_\text{orig}$ as the teacher and the hybrid model as the student. The objective can be formulated as
\begin{equation}
    \begin{aligned}
        \mathcal{L}_{\text{stage 2}} = D_\text{KL} \left(f_\text{orig}(\mathbf X) \Vert f_\text{hybrid}(\mathbf X) \right),
    \end{aligned}
\end{equation}
where $D_\text{KL}$ is KL divergence. The teacher model weights are frozen in this stage. We use 1B training data for knowledge distillation, and adopt a cosine learning rate~(LR) scheduler that decays from $\eta_\text{stage2}$ to 1e-5, where $\eta_\text{stage2}$ is determined by a separate hyperparameter search for each model size. The effectiveness of this distillation setting is validated in Appendix~\ref{sec:conversion-ablation}.

\subsection{Stage 3: Finetuning}
\label{sec:stage3}

Finally, to optimize the hybrid model's capabilities, we finetune the hybrid model with greater context length and a smaller learning rate. We use 1B training data for long-context finetuning.

\section{\ARCH: An Effective Attention-RNN Hybrid Architecture}
\label{sec:model-arch}

\ARCH is illustrated in Figure~\ref{fig:arch}. It incorporates a novel PE scheme called HyPE (described in Section~\ref{sec:hype}) and some other architectural modifications (described in Section~\ref{sec:other-arch-changes}). These architectural improvements are agnostic to the RNN mixer. Therefore, \ARCH is compatible with most modern RNNs~(see Section~\ref{sec:rnn-mixer} for details). A complete formulation of \ARCH can be found in Appendix~\ref{sec:appendix-complete-formulation}.

\subsection{HyPE: Hybrid Positional Encoding (\ding{202})}
\label{sec:hype}

In brief, HyPE applies RoPE in RNN layers and NoPE in attention layers. This scheme allows the model to combine the length generalization power of NoPE and the rich positional information of RoPE, getting the best of both worlds.

\paragraph{Motivation}

HyPE is motivated by the finding that RNNs have a limited ``receptive field'', which means they struggle to model long-context dependencies~\citep{stuffed-mamba}. This implies that in hybrid models, RNN layers primarily model short-distance dependencies while attention layers model long-distance dependencies. Therefore, when the context length exceeds the RNNs' receptive field, RNN layers are agnostic to the context length, implying that length generalization is unaffected by these layers. Consequently, the model's length generalization depends only on attention layers, which use NoPE, allowing it to generalize well beyond its training context length. In the meantime, RNN layers with RoPE provide rich positional information, allowing the model to outperform a NoPE-only model.

\paragraph{Attention Logits Scaling}

As the context length increases, the entropy of attention scores increases, resulting in poor length generalization. To mitigate this, we adopt the dynamic attention scaling from \citet{swan-gpt}, where the attention logits are scaled with a \textit{position-dependent scaling factor} $s_t$ during inference:
\begin{equation}
    \begin{aligned}
        \text{softmax}\left(\frac{s_t\mathbf q_t \mathbf K}{\sqrt{d_h}}\right) , \quad s_t &= \log_{a}(t + a),
    \end{aligned}
    \label{eq:logits-scaling}
\end{equation}
where $a$ is a hyperparameter determined after training by minimizing loss on a set of pre-training documents. The actual value of each model is reported in Appendix~\ref{sec:appendix-model-configs}. This scaling can be applied prior to the attention operator. Therefore, it has a negligible effect on the runtime. The effectiveness of this scaling mechanism is validated in Appendix~\ref{sec:appendix-attention-logits-scaling-results}

\paragraph{Conversion Details}

When applying \DISTILL to pre-trained checkpoints, attention layers are not trained/modified during stage 1. Therefore, the removal of RoPE in attention layers occurs at the start of stage 2, when we instantiate the final hybrid model.

\subsection{Other Architectural Modifications}
\label{sec:other-arch-changes}

In addition to HyPE, we make the following architectural modifications~(marked with \ding{203}, \ding{204}, and \ding{205} in Figure~\ref{fig:arch}) to further boost the performance and length generalization.

\paragraph{QK-Normalization (\ding{203})} Proposed by \citet{qknorm}, this normalizes $\mathbf q_t$ and $\mathbf k_t$:
\begin{equation}
\begin{aligned}
    \mathbf q_t = \text{Norm}(\mathbf x_t \mathbf W_q), \,\, \mathbf k_t =\text{Norm}(\mathbf x_t \mathbf W_k).
\end{aligned}
\end{equation}
This has been adopted by some open-source Transformer LLMs (e.g., Qwen3 and Gemma3~\citep{gemma3}), but is not usually used in RNN layers. However, we find that adding them in RNN layers improves the hybrid model's performance. Thus, when converting models without QK-normalization, we add QK-normalization to the RNN layer.

\paragraph{GQA to MHA (\ding{204})}

Most Transformer models employ grouped-query attention (GQA)~\citep{gqa}, where groups of attention heads share the same set of KVs, reducing KV cache size. However, RNN layers do not have a KV cache, and sharing KVs may reduce the expressivity of RNN layers. Thus, when initializing RNN layers before stage 1, we decouple KV heads by cloning the attention KV projection weights:
\begin{equation}
    \begin{aligned}
        \mathbf W_{\square}^{(i)} \leftarrow \mathbf W_{\square}^{(\lfloor i/g \rfloor)} ,\,\,  \forall i \in \{1,\cdots,n_h\}, \, \square \in  \{k,v\}
    \end{aligned}
\end{equation}
where $g$ is the query group size and $\mathbf W^{(i)}_\square$ is the KV projection weights for the $i$-th head.

\paragraph{Output Gate (\ding{205})}

Many recurrent architectures \citep{mamba2,gdn} have an \textit{output gate}, a data-dependent element-wise gating mechanism prior to the output projection:
\begin{equation}
    \begin{aligned}
        \mathbf o_t &= \text{Mixer}(\mathbf x_t), \quad \mathbf z_t= \sigma(\mathbf x_t \mathbf W_z), \\
        \mathbf y_t &= \left(\text{Norm}(\mathbf o_t) \odot \mathbf  z_t \right)\mathbf W_o^\top,
    \end{aligned}
\end{equation}
where $\sigma$ is an activation function, and $\mathbf W_z \in \mathbb R^{d\times d}$ is learnable parameters. We found that adding this component during conversion gives consistent performance gains with little increase in inference costs. Hence, during initialization, we add this mechanism by randomly initializing $\mathbf W_z$.

\citet{gated-attn} have shown that adding an output gate to softmax attention improves model quality and length generalization. Thus, we also add a randomly initialized output gate to attention layers, but at the start of stage 2 instead of stage 1, since attention layers are not trained in stage 1.

\paragraph{Increased Model Size}

Due to the introduction of \ding{204} and \ding{205}, \ARCH is roughly 10\% larger than the model it is distilled from. However, according to \citet{cost-optimal-gqa}, increasing model size while reducing the KV size is more cost-effective in long-context scenarios. \ARCH is much more efficient than the base model, due to a much smaller KV cache despite having slightly more parameters.

\subsection{RNN Mixer}
\label{sec:rnn-mixer}

\ARCH is agnostic to the RNN mixer as long as it takes QKV as the input. Thus, \ARCH can flexibly adopt any of the modern RNN mixers, including Lightning attention~\citep{lightning-attn}, Mamba2~\citep{mamba2}, GLA~\citep{gla}, GDN~\citep{gdn}, and RWKV-7~\citep{rwkv7} (see Appendix~\ref{sec:appendix-compatible-rnn-mixers} for which RNN mixers are compatible).
We tried to convert Qwen3-1.7B with each mixer and concluded that Lightning Attention provides the best balance between CSR and length generalization. The ablation results are reported in Section~\ref{sec:training-from-scratch-results}.

\begin{table*}[!t]
    \setlength{\tabcolsep}{3pt}
    \centering
    \footnotesize
    \caption{Long-context recall performance of \ARCH + \DISTILL versus state-of-the-art hybrid models that are distilled from pre-trained Transformer models. Qwen3 is evaluated with YaRN, as suggested by its authors. Best scores are bolded.}
    \begin{tabular}{p{3.8cm} cc|cccc|cccc|cccc}
        \toprule
            & &
            & \multicolumn{4}{c|}{\textbf{NIAH-Single-1}} 
            & \multicolumn{4}{c|}{\textbf{NIAH-Single-2}} 
            & \multicolumn{4}{c }{\textbf{NIAH-Single-3}}\\
        \cmidrule(lr){4-7} 
        \cmidrule(lr){8-11} 
        \cmidrule(lr){12-15}
        \textbf{Model} & \textbf{Param} & \textbf{Token} 
            & \textbf{32K} & \textbf{64K} & \textbf{128K} & \textbf{256K}
            & \textbf{32K} & \textbf{64K} & \textbf{128K} & \textbf{256K} 
            & \textbf{32K} & \textbf{64K} & \textbf{128K} & \textbf{256K} \\
        \midrule
        \rowcolor{gray!10}
        \textit{Qwen3 (teacher, no RNNs)} 
            & 1.7B & -
            & 100   & 100   & 96.4   & 17.0
            & 100   & 98.8   & 24.8   & 19.2
            & 100   & 98.4   & 14.8   & 19.0 \\
        \midrule
        % \midrule
        Jet-Nemotron \citep{jet-nemotron}
            & \textbf{2B} & 400B		
            & \textbf{99.8}  & 56.0  & 0.0   & 0.0 
            & 94.2  & 65.0  & 0.0   & 0.0
            & 84.0  & 15.4  & 0.0   & 0.0\\
        % \midrule
        KL-LS (GDN) \citep{kl-guided-layer-selection}
            & 3B & 25B 
            & \textbf{99.8}  & 99.4  & 68.4  & 14.8 
            & \textbf{99.4}   & 49.6     & 28.2     & 10.4
            & \textbf{99.0}   & 51.0     & 24.8     & 11.0 \\
            
        \rowcolor{orange!10} 
        \ARCH + \DISTILL (ours) 
            & \textbf{2B} & \textbf{2.3B}
            & \textbf{99.8}  & \textbf{99.6}  & \textbf{99.8} & \textbf{99.8}
            & 95.2  & \textbf{99.6}  & \textbf{97.8} & \textbf{86.2}
            & 87.2  & \textbf{72.6}  & \textbf{44.8} & \textbf{48.8} \\
        \bottomrule
    \end{tabular}
    \label{tab:results-hybrid-models}
\end{table*}

\section{Experiments}
\label{sec:experiments}

We first describe our experimental setup~(Section~\ref{sec:experimental-setup}).
Then, we compare \ARCH + \DISTILL against Qwen3 and state-of-the-art hybrids that are also converted from pre-trained models~(Section~\ref{sec:main-results}).
Then, we verify the effectiveness of various design choices in \ARCH~(Section~\ref{sec:training-from-scratch-results}). Afterwards, we present ablation studies for \DISTILL's architectural modifications~(Section~\ref{sec:arch-ablation-converting-from-qwen3}) and attention layer selection method~(Section~\ref{sec:layer-selection-results}).
Finally, we analyze the inference efficiency of \ARCH~(Section~\ref{sec:efficiency-results}).

\subsection{Experimental Setup}
\label{sec:experimental-setup}

\paragraph{Models}

We apply \DISTILL to the 1.7B, 4B, and 8B models of Qwen3~\citep{qwen3}, which is one of the most widely-used open-source language model series.

\paragraph{Training Configurations}

In \DISTILL, we use FineWeb-edu~\citep{fineweb-edu} for training. It is a popular open-source, high-quality Internet-scale pre-training corpus. All data are randomly sampled from the 10B subset. The concrete hyperparameters that we use for each stage in \DISTILL are reported in Appendix~\ref{sec:appendix-training-configs}.

\paragraph{Evaluation}

We mainly evaluate CSR and long-context recall performance. For CSR, we use a suite of zero-shot downstream tasks that are common in related literature. To measure long-context performance, we report accuracy on NIAH\footnote{By default, NIAH refers to the average of NIAH-Single-1, NIAH-Single-2, and NIAH-Single-3 from RULER.}. More details are given in Appendix~\ref{sec:appendix-evaluation-details}.

\paragraph{Evaluation Data for Layer Selection}

Our layer selection method relies on measuring the performance change in CSR and recall~(see Eq.~(\ref{eq:layer-importance})). Inspired by \citet{jet-nemotron}, we use the normalized accuracy on HellaSwag~\citep{hellaswag}, ARC-Easy, and ARC-Challenge~\citep{arc} as the CSR performance, the average score on SQuAD~\citep{squad}, FDA~\citep{fda}, and SWDE~\citep{swde} as the recall performance.

\paragraph{Efficiency Measurement}

All efficiency measurements are conducted on servers with a single NVIDIA A800 GPU, using PyTorch version 2.9.1 and CUDA version 12.4. Softmax attention is implemented with Flash-Attention-2~\citep{flashattention2}, version 2.8.3. Mamba2 is implemented its official CUDA kernel, version 2.3.0. Other RNN mixers are implemented with Triton kernels from Flash-Linear-Attention~\citep{flash-linear-attention}, version 0.4.1. Batch size is set to 1 for all models to ensure fair comparison. 

\begin{figure}[!t]
    \centering
    \includegraphics[width=\linewidth]{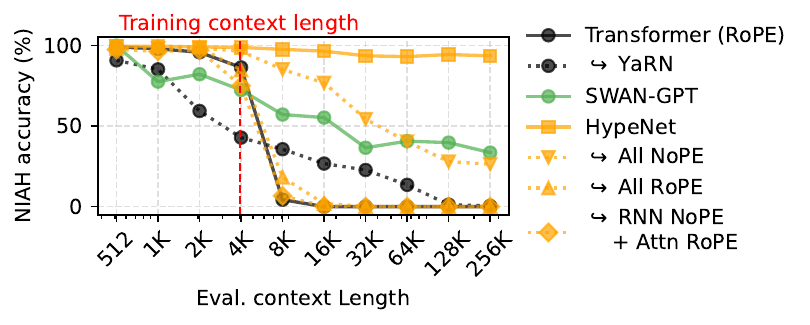}
    \caption{NIAH scores of \ARCH variants based on different position encodings, as a function of context length. The models are trained from scratch with 20B tokens and 500M parameters.}
    \label{fig:result_pos_enc_from_scratch}
\end{figure}

\begin{figure}[!t]
    \centering
    \includegraphics[width=\linewidth]{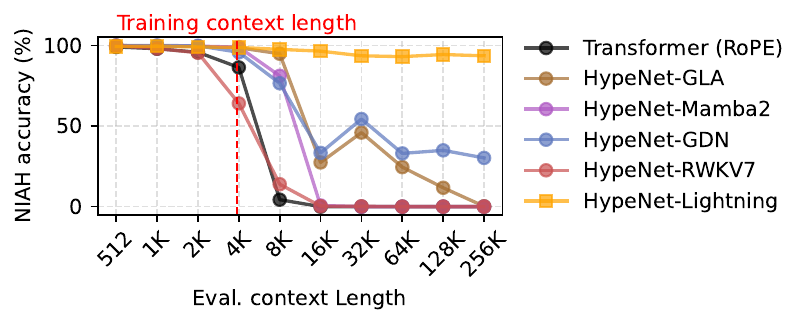}
    \caption{NIAH scores of \ARCH variants based on different RNN mixers, as a function of context length. The models are trained from scratch with 20B tokens and 500M parameters.}
    \label{fig:result_rnn_mixer_from_scratch}
\end{figure}

\subsection{Main Results: Distilling from Qwen3}
\label{sec:main-results}
% \label{sec:results}

Figure~\ref{fig:performance-vs-cost} shows the CSR performance and efficiency of \ARCH compared to the Qwen3 series, and Table~\ref{tab:results-hybrid-models} reports the long-context recall performance. 
Also in Table~\ref{tab:results-hybrid-models}, \ARCH + \DISTILL is compared against recently released state-of-the-art hybrid models that are distilled from pre-trained Transformer models. 

\paragraph{Takeaway 1}
Under 128K context length, \ARCH is much more efficient than Qwen3 in terms of memory and throughput due to the reduced number of attention layers, and this tradeoff advantage increases with the context length.

\paragraph{Takeaway 2}
Compared to state-of-the-art Transformer-to-hybrid methods, \ARCH + \DISTILL achieve superior long-context performance, despite using fewer training tokens, training with only open-source data, and being smaller than KL-LS~(GDN).

\begin{table*}[!t]
    \centering
    \footnotesize
    \caption{Ablation experiment results for various architectural choices in \ARCH-2B, converted from Qwen3-1.7B.}
    \begin{tabular}{lc |cccccc}
        \toprule
        & & \multicolumn{6}{c}{\textbf{Needle-in-a-Haystack}} \\
        \cmidrule(lr){3-8}
        \textbf{Model} & \textbf{CSR} & \textbf{4K} & \textbf{8K} & \textbf{16K} & \textbf{32K} & \textbf{64K} & \textbf{128K} \\
        \midrule
        \rowcolor{orange!10} 
        \ARCH       
            & \textbf{55.9} & \textbf{95.9} & 94.9 & \textbf{90.3} & \textbf{94.1} & \textbf{90.6} & 79.9 \\
        \midrule
        $\hookrightarrow$ w/o RNN RoPE (\ding{202})
            & 53.8 & 82.3 & 82.7 & 79.1 & 76.1 & 72.4 & 47.9 \\
        $\hookrightarrow$ w/ attention RoPE (\ding{202})
            & 55.8 & 95.3 & \textbf{95.3} & 87.0 & 67.1 & 37.2 & 19.7 \\
        \midrule
        $\hookrightarrow$ w/o RNN QK-norm (\ding{203})
            & 55.3 & 91.7 & 92.3 & 89.1 & 73.9 & 53.5 & 17.3 \\
        \midrule
        $\hookrightarrow$ w/o RNN GQA to MHA (\ding{204})    
            & 55.8 & 89.7 & 90.0 & 87.9 & 89.5 & 88.9 & \textbf{83.5} \\
        % $\hookrightarrow$ w/o RNN short conv. 
        %     & \multicolumn{7}{c}{Failed to converge} \\
        \midrule
        $\hookrightarrow$ w/o RNN output gate (\ding{205})
            & 55.6 & 91.1 & 89.3 & 84.6 & 84.9 & 81.3 & 74.5 \\
        $\hookrightarrow$ w/o attention output gate (\ding{205})
            & 55.4 & 95.5 & 93.3 & 88.2 & 92.5 & 87.3 & 80.9 \\
        \bottomrule
    \end{tabular}
    \label{tab:arch-ablation-results}
\end{table*}

\begin{table*}[!t]
    \centering
    \footnotesize
    \caption{Comparison of different \textit{attention layer selection} methods on CSR and NIAH tasks. All models are converted from Qwen3 with \DISTILL, but use different layer selection methods. The best scores are bolded.}
    % \begin{tabular}{l c c c| cccccc}
    \begin{tabular}{l c| cccccc}
        \toprule
            % & \textbf{Model} 
            % & \textbf{Train} 
            &
            & \multicolumn{6}{c}{\textbf{Needle-in-a-Haystack}} \\
        % \cmidrule(lr){5-10}
        \cmidrule(lr){3-8}
        \textbf{Model} 
            % & \textbf{Size} 
            % & \textbf{Tokens} 
            & \textbf{CSR} 
            % & \textbf{4K} 
            & \textbf{8K} 
            & \textbf{16K} 
            & \textbf{32K} 
            & \textbf{64K} 
            & \textbf{128K} 
            & \textbf{256K} \\
        \midrule
        \rowcolor{gray!10}
        \textit{Qwen3-1.7B (teacher, no RNNs)} 
            % & 1.7B
            % & -
            & 58.5 
            % & 100 
            & 99.7 
            & 99.9 
            & 99.9
            & 99.5 
            & 38.6
            & 18.4 \\ 
        \midrule
        \rowcolor{orange!10}
        \DISTILL (ours)
            % & 2B
            % & 2.3B 
            & \textbf{55.9}
            & \textbf{94.9} & \textbf{90.3} & \textbf{94.1} & \textbf{90.6} & \textbf{79.9} & \textbf{74.3} \\
        Jet-Nemotron-2B~\citep{jet-nemotron}
            % & 2B
            % & 2.3B 
            & 55.0 
            % & \textbf{95.3} 
            & 88.7 
            & 70.1 
            & 70.3 
            & 61.9 
            & 63.7 
            & 56.2 \\
        KL-LS~\citep{kl-guided-layer-selection}
            % & 2B
            % & 2.3B 
            & 55.3 
            % & 89.1 
            & 85.7 
            & 78.4 
            & 72.8 
            & 68.9 
            & 58.3
            & 44.3 \\
        Evenly distribute attn. layers
            % & 2B
            % & 2.3B 
            & 54.0 
            % & 77.6 
            & 78.1 
            & 77.8 
            & 68.2 
            & 73.5 
            & 61.9
            & 50.9\\
        Evenly distribute attn. layers in the latter half
            % & 2B
            % & 2.3B 
            & 55.8
            % & 54.5 
            & 42.5 
            & 39.6 
            & 50.5 
            & 41.2 
            & 39.2
            & 40.4 \\
        \midrule
        \textit{RADLADS (RNN-only) \citep{radlads}} 
            % & 1.7B
            % & -
            & 56.0 
            & 64.1 
            & 16.4 
            & 2.0 & 0.0 & 0.0 & 0.0 \\
        \bottomrule
    \end{tabular}
    \label{tab:results-attn-layer-selection}
\end{table*}

\subsection{\ARCH Ablations: Training From Scratch}
\label{sec:training-from-scratch-results}

To validate the effectiveness of \ARCH, we pre-train 500M \ARCH variants from scratch with 20B tokens and compare them against common baselines. The experimental details are reported in Appendix~\ref{sec:appendix-training-from-scratch-configs}.

\paragraph{Position Encoding}

We compare HyPE against ordinary Transformer with RoPE and SWAN-GPT~\citep{swan-gpt}, which is an architecture with a similar PE but is not a hybrid model. We also compare with \ARCH variants without HyPE (i.e., all RoPE, all NoPE, or attention RoPE + RNN NoPE). The result, reported in Figure~\ref{fig:result_pos_enc_from_scratch}, demonstrates that HyPE's length generalization abilities outperform existing PE by a large margin. Notably, we find that, compared to conversion from pre-trained checkpoints, training HyPE from scratch achieves even better length generalization (having 93.5\% NIAH accuracy on 64$\times$ the training context length), demonstrating the great potential of HyPE.

\paragraph{Different RNN Mixers}

Moreover, we also compare the performance of incorporating different RNN mixers (those mentioned in Section~\ref{sec:rnn-mixer}), and report the results in Figure~\ref{fig:result_rnn_mixer_from_scratch}. Perhaps surprisingly, Lightning Attention outperforms more recent RNN variants in terms of length generalization despite having a simpler update rule. One possible explanation is that Lightning Attention employs \textit{data-independent} forget gates. In contrast, the other RNN mixers have \textit{data-dependent} forget gates, which may result in poor length generalization, as shown by \citet{stuffed-mamba}.

\paragraph{Takeaway}
The incorporation of HyPE and Lightning Attention is both essential for achieving the exceptional length generalization of \ARCH.

\subsection{\DISTILL Ablations: Architectural Modifications}
\label{sec:arch-ablation-converting-from-qwen3}

This section validates the effectiveness of various architectural modifications of \DISTILL~(those marked with \ding{202}, \ding{203}, \ding{204}, and \ding{205} in Figure~\ref{fig:arch}).
Table~\ref{tab:arch-ablation-results} reports the ablation results when converting Qwen3-1.7B, and it shows that our architectural modifications provide effective gains in CSR and NIAH performance, considerably outperforming common approaches in training hybrid architectures.

\subsection{\DISTILL Ablations: Attention Layer Selection}
\label{sec:layer-selection-results}

Here, we compare our proposed layer selection method~(described in Section~\ref{sec:attn-layer-selection}) with two state-of-the-art approaches for determining layer importance, Jet-Nemotron~\citep{jet-nemotron} and KL-LS~\citep{kl-guided-layer-selection}, as well as some naive baselines that evenly distribute attention layers. We do not run the entire distillation procedures of Jet-Nemotron or KL-LS, which involve training on much more data. Our comparison is performed by replacing our attention layer selection method in \DISTILL with these previous methods. The result is reported in Table~\ref{tab:results-attn-layer-selection}, and it shows that our selection method achieves a better overall performance in terms of CSR and recall.

\subsection{Efficiency Results}
\label{sec:efficiency-results}

\begin{figure}[!h]
    \centering
    \includegraphics[width=0.9\linewidth]{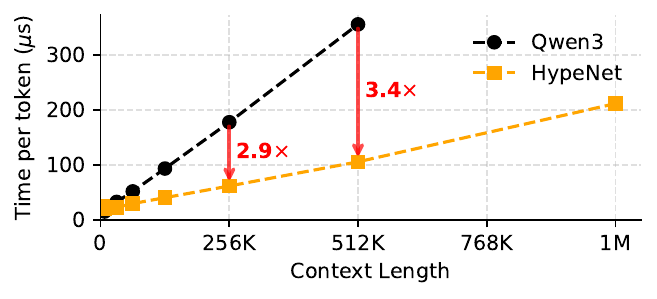}
    \caption{The prefilling time of \ARCH versus Qwen3-1.7B, across different context lengths.}
    \label{fig:prefill-speed}
\end{figure}

Figure~\ref{fig:performance-vs-cost}~(center) shows the throughput of \ARCH models of different sizes~(2B, 5B, and 9B) at 128K context length, and Figure~\ref{fig:performance-vs-cost}~(right) shows the time per output token~(TPOT) across different context lengths. Figure~\ref{fig:prefill-speed} shows the prefill speed results.
We also provide a comparison of the runtime of various RNN mixers in Appendix~\ref{sec:appendix-rnn-mixer-efficiency}. In brief, \ARCH achieves up to 3.0$\times$ decoding speedup and 3.4$\times$ prefilling speedup on 512K context length, before Qwen3-1.7B runs out of GPU memory on 1M context length.

\section{Conclusion}

We have proposed \DISTILL, a novel distillation procedure for converting pre-trained Transformer models into RNN-attention hybrid architectures with less than 3B tokens.
We also proposed \ARCH, a hybrid architecture based on a novel PE scheme called HyPE, and it achieves superior length generalization. Applying our methods to Qwen3 produces a series of hybrid models with much better performance-throughput tradeoff and memory-efficiency on long-context scenarios. We believe that our work is valuable for research in cost-efficient long-context LLMs, which enables many useful applications such as long-horizon reasoning and agentic behaviors. Our work also fosters research in novel LLM architectures by making it cheaper to empirically validate hybrid architectures at scale.

\section*{Limitations}

Our hybrid models are obtained through a conversion process trained on the FineWeb-Edu corpus, which primarily consists of pre-training-style data. As a result, instruction-following and alignment behaviors of the pre-training model introduced by post-training may be diminished by our conversion process. However, this is a common shortcoming of all existing distillation methods for converting into hybrid architectures. How to efficiently recover the base models' capabilities remains an open question. 

Moreover, our conversion protocol is designed specifically for Transformer-based architectures. Hence, its applicability to other model architectures requires further investigation, although the vast majority of publicly available LLMs are Transformer-based.

\section*{Impact Statement}

This paper presents work whose goal is to advance the field of machine learning. There are many potential societal consequences of our work, none of which we feel must be specifically highlighted here.

% \bibliography{example_paper}
\bibliography{custom}
\bibliographystyle{icml2026}

%%%%%%%%%%%%%%%%%%%%%%%%%%%%%%%%%%%%%%%%%%%%%%%%%%%%%%%%%%%%%%%%%%%%%%%%%%%%%%%
%%%%%%%%%%%%%%%%%%%%%%%%%%%%%%%%%%%%%%%%%%%%%%%%%%%%%%%%%%%%%%%%%%%%%%%%%%%%%%%
% APPENDIX
%%%%%%%%%%%%%%%%%%%%%%%%%%%%%%%%%%%%%%%%%%%%%%%%%%%%%%%%%%%%%%%%%%%%%%%%%%%%%%%
%%%%%%%%%%%%%%%%%%%%%%%%%%%%%%%%%%%%%%%%%%%%%%%%%%%%%%%%%%%%%%%%%%%%%%%%%%%%%%%
\newpage
\appendix
\onecolumn

\section{Complete Formulation of \ARCH}
\label{sec:appendix-complete-formulation}

Here, we present a complete formulation of \ARCH for clarity. Recall that the model consists of a stack of $L$ layers that consists of a token mixer and MLP:
\begin{equation}
    \begin{aligned}
        \mathbf H^{(l)} &= \text{Mixer}^{(l)}\left( \text{Norm} \left(\mathbf X^{(l-1)} \right)\right) + \mathbf X^{(l-1)} \in \mathbb R^{T\times d} \\
        \mathbf X^{(l)} &= \text{MLP}^{(l)} \left(\text{Norm} \left(\mathbf H^{(l)} \right)\right) +\mathbf H^{(l)} \in \mathbb R^{T\times d}
    \end{aligned}
\end{equation}
where $l\in\{1,\cdots,L\}$ is the layer index and $\text{Norm}(\cdot)$ represents an RMSNorm~\citep{rmsnorm}. Then, each mixer is either an attention layer $\text{ATTN}(\cdot)$ or an RNN layer $\text{RNN}(\cdot)$, specified by an attention index set $\mathcal I_\text{attn}$:
\begin{equation}
    \begin{aligned}
        \text{Mixer}^{(l)} = \begin{cases}
            \text{ATTN}^{(l)} & \text{if } l \in \mathcal I_\text{attn} \\
            \text{RNN}^{(l)} & \text{otherwise}
        \end{cases}
    \end{aligned}
\end{equation}
Since the MLP layer is exactly the same as the one in the base model, we omit its formulation. Each attention layer and RNN layer consists of $n_h$ heads, that are identical (except for the KV sharing mechanism in GQA). Thus, in the following formulations, we omit the head index and only give the formulation for a single head for simplicity. The output of the layer is the sum of the outputs of all heads.

\paragraph{Attention Layers}
Each attention layer can be written as follows:
\begin{equation}
    \begin{aligned}
        \mathbf q_t &= \mathbf x_t \mathbf W_q \in\mathbb R^{1\times d_h}\\
        \mathbf k_t &= \mathbf x_t \mathbf W_k \in\mathbb R^{1\times d_h}\\
        \mathbf v_t &= \mathbf x_t \mathbf W_v \in\mathbb R^{1\times d_h}\\
        \mathbf{\tilde q}_t &= \frac{s_t \mathbf{q}_t}{\sqrt{d_h}}\in\mathbb R^{1\times d_h}, \quad s_t = \log_a(t+a) \in \mathbb R \\
        \mathbf o_t &= \sum_{i=1}^t \frac{\exp \left(\mathbf{\tilde q}_t \mathbf k_i^\top \right) \mathbf v_i}{\sum_{j=1}^t \exp \left(\mathbf {\tilde q}_j\mathbf k_j^\top\right)} \in\mathbb R^{1\times d_h}\\
        \mathbf z_t &= \text{sigmoid}(\mathbf x_t\mathbf W_z)  \in\mathbb R^{1\times d_h}\\
        \mathbf y_t &= \left(\text{Norm}(\mathbf o_t) \odot \mathbf z_t \right) \mathbf W_o^\top  \in\mathbb R^{1\times d}
    \end{aligned}
\end{equation}
where $\mathbf W_q,\mathbf W_k,\mathbf W_v,\mathbf W_o,\mathbf W_z \in \mathbb R^{d\times d_h}$ are learnable parameters, and $s_t$ is the position-dependent scaling factor, and $a$ is a hyperparameter. Depending on the base model, there may be a QK-norm in attention layers.

\paragraph{RNN Layers}
Each RNN layer can be written as follows:
\begin{equation}
    \begin{aligned}
        \mathbf q_t &= \text{Norm}\left(\mathbf x_t \mathbf W_q \right)\in\mathbb R^{1\times d_h}\\
        \mathbf k_t &= \text{Norm}\left(\mathbf x_t \mathbf W_k \right)\in\mathbb R^{1\times d_h} \\
        \mathbf v_t &= \mathbf x_t \mathbf W_v \in\mathbb R^{1\times d_h}\\
        \mathbf{\tilde q}_t &= \text{RoPE}_{t}(\mathbf q_t) \in\mathbb R^{1\times d_h}\\
        \mathbf{\tilde k}_t &= \frac{\text{RoPE}_{t}(\mathbf k_t)}{\sqrt {d_h}} \in\mathbb R^{1\times d_h}\\
        \mathbf S_t &= \mathbf S_{t-1} \gamma  + \mathbf {\tilde k}_t ^\top \mathbf v_t \in\mathbb R^{d_h \times d_h}\\
        \mathbf o_t &= \mathbf {\tilde q}_{t} \mathbf S_t \in\mathbb R^{1\times d_h}\\
        \mathbf z_t &= \text{sigmoid}(\mathbf x_t\mathbf W_z) \in\mathbb R^{1\times d_h}\\
        \mathbf y_t &= \left(\text{Norm}(\mathbf o_t) \odot \mathbf z_t \right) \mathbf W_o^\top \in\mathbb R^{1\times d}
    \end{aligned}
\end{equation}
where $\mathbf W_q,\mathbf W_k,\mathbf W_v,\mathbf W_o,\mathbf W_z \in \mathbb R^{d\times d_h}$ are learnable parameters, and $\gamma$ is the head-specific slope rate of Lightning Attention~\citep{lightning-attn}, which is a \textit{data-independent forget gate}. $\text{RoPE}_t$ is the rotational matrix of RoPE~\citep{rope} for position $t$.

\paragraph{Forget Gate}

The forget gate of Lightning Attention in \ARCH is defined as:
\begin{equation}
    \gamma_h = \exp \left(-2^{-8h/H}\right) \in \mathbb (0,1)
\end{equation}
where $h\in\{1,\cdots,H\}$ is the head index and $H$ is the number of heads. Notably, we do not rescale this value with a layer-specific factor as in the original implementation, because our preliminary results show that it does not yield performance gains in a hybrid model. The $\gamma_h$ values for each head when $H=32$ is:
\begin{footnotesize}
\begin{verbatim}
  0.4313237   0.4930687   0.5517813   0.60653067  0.6567524   0.7021885   0.74281985
  0.7788008   0.81040263  0.83796686  0.86186993  0.8824969   0.9002237   0.91540533  
  0.9283695   0.9394131   0.94880116  0.95676816  0.96351933  0.9692332   0.97406423 
  0.97814524  0.9815902   0.9844964   0.98694694  0.98901224  0.99075234  0.99221796  
  0.993452    0.994491    0.99536544  0.9961014                                                 
\end{verbatim}
\end{footnotesize}

\section{\DISTILL Training Configurations}
\label{sec:appendix-training-configs}

Table~\ref{tab:hyperparameters-distillation} reports the hyperparameters used for each stage in our conversion procedure, \DISTILL. By default, we use AdamW optimizer with beta values of $(0.9, 0.95)$ and without weight decay. Each stage use an LR linear warmup from 0 to maximum LR, consisting of 50 steps. We train all models with BFloat16 precision.

\begin{table*}[!ht]
    \centering
    \footnotesize
    \caption{Hyperparameters for each training stage in \DISTILL. $\eta_\text{stage2}$ is the a hyperparameter that depends on the model (reported in Table~\ref{tab:hyperparameters-arch}).}
    \begin{tabular}{l|cccccc}
        \toprule
        \textbf{Stage} & \textbf{Tokens} & \textbf{LR Scheduler} & \textbf{LR} & \textbf{Context len.} & \textbf{Batch size} & \textbf{Train steps}\\
        \midrule
        1 & 320M & Cosine & 1e-3 $\rightarrow$ 1e-5 & 512 & 32 & 20K \\
        2 & 1B   & Cosine   & $\eta_\text{stage2} \rightarrow$ 1e-5 & 512 & 96 & 20K \\
        3 & 1B   & Constant   & 1e-5 & 16K & 128 & 500 \\
        \bottomrule
    \end{tabular}
    \label{tab:hyperparameters-distillation}
\end{table*}

\section{\ARCH Model Configurations}
\label{sec:appendix-model-configs}

Table~\ref{tab:hyperparameters-arch} reports the configuration of each model in this study. We also report the actual indices of the attention layers for each \ARCH model in Table~\ref{tab:attn-layer-indices}.

\begin{table*}[!ht]
    \centering
    \footnotesize
    \caption{Hyperparameters of various \ARCH models.}
    \begin{tabular}{l|ccc}
        \toprule
        \textbf{Hyperparameter} & \textbf{\ARCH-2B} & \textbf{\ARCH-5B} & \textbf{\ARCH-9B} \\
        \midrule
        Vocab size          & 151936 & 151936 & 151936\\
        Layers              & 28    & 36    & 36    \\
        Hidden size         & 2048  & 2560  & 4096   \\
        RNN layers          & 7     & 8     & 8      \\
        Attn. layers        & 21    & 24    & 24     \\
        FFN width           & 6144  & 9728  & 12288  \\
        Attention heads     & 16    & 32    & 32     \\
        Attention KV heads  & 8     & 8     & 8      \\
        RNN heads           & 16    & 32    & 32     \\
        Tie embeddings      & Yes   & Yes   & Yes    \\
        RoPE theta          & 1M    & 1M    & 1M     \\
        RoPE scaling        & None  & None  & None   \\
        \midrule
        $a$ (in Eq.~(\ref{eq:logits-scaling})) & 500 & 600 & 900 \\
        $\eta_\text{stage2}$ (see Table~\ref{tab:hyperparameters-distillation}) & 1e-4 & 5e-5 & 3e-5 \\
        \bottomrule
    \end{tabular}
    \label{tab:hyperparameters-arch}
\end{table*}

% \begin{table*}[!ht]
%     \centering
%     \footnotesize
%     \caption{Hyperparameters of various \ARCH models.}
%     \begin{tabular}{l|cccc}
%         \toprule
%         \textbf{Hyperparameter} & \textbf{\ARCH-2B} & \textbf{\ARCH-5B} & \textbf{\ARCH-9B} & \textbf{\ARCH-MiniCPM-9B} \\
%         \midrule
%         Vocab size          & 151936 & 151936 & 151936 & 73448\\
%         Layers              & 28    & 36    & 36    & 32 \\
%         Hidden size         & 2048  & 2560  & 4096  & 3072 \\
%         RNN layers          & 7     & 8     & 8     & 9 \\
%         Attn. layers        & 21    & 24    & 24    & 27 \\
%         FFN width           & 6144  & 9728  & 12288 & 16384 \\
%         Attention heads     & 16    & 32    & 32    & 32 \\
%         Attention KV heads  & 8     & 8     & 8     & 2 \\
%         RNN heads           & 16    & 32    & 32    & 32 \\
%         Tie embeddings      & Yes   & Yes   & Yes   & No \\
%         RoPE theta          & 1M    & 1M    & 1M    & 10K \\
%         RoPE scaling        & None  & None  & None  & LongRoPE \\
%         \midrule
%         $a$ (in Eq.~(\ref{eq:logits-scaling})) & 500 & 600 & 900 & 500 \\
%         $\eta_\text{stage2}$ (see Table~\ref{tab:hyperparameters-distillation}) & 1e-4 & 5e-5 & 3e-5 & 1e-5 \\
%         \bottomrule
%     \end{tabular}
%     \label{tab:hyperparameters-arch}
% \end{table*}

\section{Addition Notes on the Model Architecture}
\label{sec:appendix-additional-notes-on-arch}

\begin{table*}[!ht]
    \centering
    \footnotesize
    \caption{Layer selection results. Here are the attention layers indices sorted by importances score computed using differenct layer selection methods. The top-$k$ attention layers that are kept in the final model are highlighted with a box. The \textcolor{red}{red} indices in the box indicate layers that are not selected by our approach.}
    \begin{tabular}{p{3.8cm}|p{12cm}}
        \toprule
        \textbf{Method} & \textbf{Layer indices (most important $\rightarrow$ least important)} \\
        \midrule
        \multicolumn{2}{c}{\textit{Qwen3-1.7B}} \\
        \midrule
        HALO (ours) & \boxed{3, 21, 2, 9, 25, 6, 8,} 19, 16, 24, 12, 26, 23, 11, 27, 14, 18, 4, 7, 17, 13, 15, 20, 10, 22, 1, 0, 5 \\
        \midrule
        Jet-Nemotron~\citep{jet-nemotron} & \boxed{\textcolor{red}{0}, 21, 25, \textcolor{red}{19}, 6, \textcolor{red}{11}, 9,} 24, 12, 2, 26, 16, 17, 23, 18, 4, 7, 3, 14, 20, 1, 27, 10, 13, 8, 22, 15, 5 \\
        \midrule
        KL-guided layer selection~\citep{kl-guided-layer-selection} & \boxed{21, \textcolor{red}{16}, 25, \textcolor{red}{24}, \textcolor{red}{0}, \textcolor{red}{18}, \textcolor{red}{19},} 20, 8, 1, 2, 11, 12, 26, 13, 17, 14, 15, 10, 9, 22, 23, 6, 7, 4, 3, 27, 5 \\
        \midrule
        \multicolumn{2}{c}{\textit{Qwen3-4B}}\\
        \midrule
        HALO (ours) & \boxed{0, 7, 1, 33, 24, 15, 34, 22,} 14, 31, 5, 21, 23, 16, 20, 2, 18, 19, 32, 27, 13, 25, 30, 6, 29, 17, 11, 35, 8, 12, 9, 10, 26, 28, 4, 3\\
        % \midrule
        % Jet-Nemotron & \underline{7, 24, 0, 22, 33, 15, 34, 14}, 21, 5, 19, 16, 20, 18, 31, 23, 1, 8, 32, 13, 27, 2, 25, 35, 17, 6, 29, 11, 26, 12, 30, 10, 9, 28, 4, 3\\
        % \midrule
        % KL-guided layer selection & \underline{22, 24, 20, 23, 13, 26, 27, 0} 19, 5, 17, 16, 11, 21, 4, 15, 18, 14, 9, 6, 2, 25, 12, 3, 10, 8, 1, 7 \\
        \midrule
        \multicolumn{2}{c}{\textit{Qwen3-8B}}\\
        \midrule
        HALO (ours) & \boxed{10, 6, 7, 24, 33, 2, 4, 1, 34,} 22, 13, 26, 35, 20, 31, 15, 9, 29, 14, 5, 3, 17, 23, 28, 30, 21, 25, 18, 8, 11, 32, 12, 0, 19, 27, 16 \\
        % \midrule
        % Jet-Nemotron & 0, 24, 33, 7, 22, 2, 13, 26, 15, 5, 20, 14, 29, 17, 31, 21, 1, 34, 23, 3, 4, 9, 6, 32, 25, 18, 35, 28, 11, 30, 8, 10, 12, 19, 27, 16 \\
        % \midrule
        % KL-guided layer selection \\
        \bottomrule
    \end{tabular}
    \label{tab:attn-layer-indices}
\end{table*}

\paragraph{Short Convolution}

Many recent RNNs~\citep{mamba2,gdn,jet-nemotron} incorporate a ``short convolution'' layer, which is a per-channel 1D convolutional layer with a small kernel size (typically from 2 to 4). Most transformer models do not have this layer. Consistent with \citet{radlads}, we found that adding this component through post-training does not provide performance gains for the 8B model and even failed to converge when applied to the 1.7B model. Moreover, short convolutional layers require another dedicated CUDA kernel and more implementation overhead. Thus, we do not incorporate short convolutional layers in \ARCH.

\section{Computational Cost of Each Stage}
\label{sec:appendix-computational-cost-of-each-stage}

\begin{table*}[!ht]
    \centering
    \footnotesize
    \caption{The number of FLOPs and training/inference tokens required by each stage in \DISTILL, applied to Qwen3-1.7B. Layer selection stage spends fewer FLOPs per token because it performs only inference and does not require backward passes. Stage 3 has greater FLOPs per token because it uses a greater context length. $^\ast$ indicates inference tokens while other entries are training tokens.}
    \begin{tabular}{l|cccc}
        \toprule
        \textbf{Stage}   & \textbf{Tokens} & \textbf{FLOPs / token} & \textbf{FLOPs} & \textbf{GPU hours (A800)} \\
        \midrule
        Stage 1         & 320M          & 4.15B & 2.7e18 & 10.0 \\
        Layer selection & 234M$^\ast$   & 1.38B & 6.5e17 & N/A \\
        Stage 2         & 1B            & 4.15B & 8.3e18 & 43.4 \\
        Stage 3         & 1B            & 6.88B & 1.4e19 & 37.7 \\
        \midrule
        Total           & 2.3B          & 16.6B & 2.5e19 & 91.1 \\
        \bottomrule
    \end{tabular}
    \label{tab:computation-costs}
\end{table*}

Table~\ref{tab:computation-costs} reports the computational cost (in the number of FLOPs) of each stage in \DISTILL, our distillation process. The layer selection process requires the model to perform inference on our evaluation tasks. There tasks contain 8.36M tokens in total, and the FLOPs per token for inference is notably fewer than that of training.

\paragraph{The Number of Tokens Required by KL-Guided Layer Selection}

The number of training tokens used for the attention selection method, KL-guided layer selection~(KL-LS)~\citep{kl-guided-layer-selection}, depends on the number of layers in model. Specifically, their method requires $700 \text{M} \times L + 600\text{M}$ tokens, where $L$ is the number of layers in the base model. In the main content~(Table~\ref{tab:previous-works} and Figure~\ref{fig:pipeline}), we report the number of tokens used for converting Qwen2.5-3B into RNNs with KL-LS, which is the model used in their paper. That model has 36 layers.

\subsection{RNN Mixer Efficiency Measurement}
\label{sec:appendix-rnn-mixer-efficiency}

\begin{figure}[!ht]
    \centering
    \includegraphics[width=0.7\linewidth]{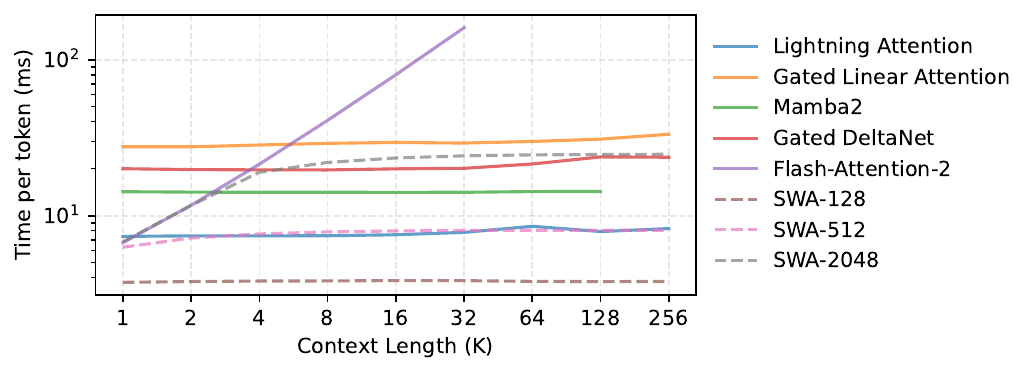}
    \caption{The inference prefilling time of various mixers as a function of context lengths, measured on one A800-80GB GPU using BFLoat16. The sliding window mixers are implemented with Flash-Attention-2, Mamba2 is implemented with its official \texttt{mamba\_ssm} library, and all other RNN mixers are taken from the widely used Flash-Linear-Attention\footnote{\url{https://www.github.com/fla-org/flash-linear-attention}}. Mamba2 ran out of CUDA Memory on 256K context length. The y-axis is on log scale.}
    \label{fig:mixer-runtime}
\end{figure}

In this section, we compare the runtime of each RNN mixer across different context lengths, measured on one NVIDIA A800-80GB GPU.
The inference throughput results is shown in Figure~\ref{fig:mixer-runtime}. ``time-dep.'' means that forget gates~(or, memory decay multiplier) depend on the current time step, while ``time-indep.'' means that forget gates are fixed. We find that Lightning Attention with data-independent forget gates is significantly faster than other RNN mixers and comparable to SWA with a 512 window size, thanks to its highly simple update rule. This result further validate the superiority of Lightning Attention on \ARCH.

\subsection{More Evaluation Details}
\label{sec:appendix-evaluation-details}

We use the popular evaluation framework, LM-Evaluation-Harness~\citep{eval-harness}, for all of our evaluation, and the version we use is 0.4.10.dev0. Before evaluation, we export each checkpoint such that it can be loaded with \texttt{AutoModelForCausalLM.from\_pretrained} with the HuggingFace transformers library. Then, we run LM-Evaluation-Harness with the HuggingFace API. We use BFloat16 during evaluation.

\paragraph{Qwen3 YaRN}

By default, we evaluate Qwen3 models without any modifications to the official model configuration file. But for long-context tasks that exceed their default maximum context length, which is 40,960 tokens, we apply YaRN method as described in the official model card adding a \texttt{"rope\_scaling"} entry in the configuration file.

\paragraph{Downstream Tasks for CSR}

The downstream tasks for measuring CSR performance are as follows:
\begin{itemize}
    \item ARC-Easy~\citep{arc}
    \item ARC-Challenge~\citep{arc}
    \item HellaSwag~\citep{hellaswag}
    \item WinoGrande~\citep{winogrande}
    \item PIQA~\citep{piqa}
    \item LAMBADA~\citep{lambada}
    \item MMLU~\citep{mmlu}
\end{itemize}
We always use normalized accuracy by default, which more common according to the authors of LM-Evaluation-Harness.

\section{More Experimental Results}
\label{sec:appendix-more-results}

\subsection{Attention Logits Scaling Validation}
\label{sec:appendix-attention-logits-scaling-results}

\begin{figure}[!t]
    \centering
    \includegraphics[width=0.5\linewidth]{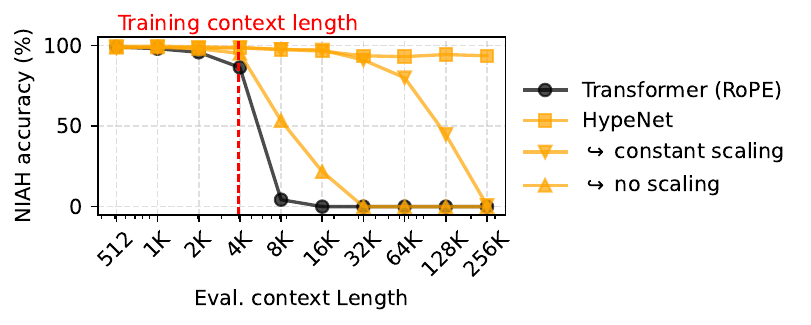}
    \caption{Results for validating attention logits scaling~(see Eq.~\ref{eq:logits-scaling}). The plot shows the NIAH performance of \ARCH without attention logits scaling, \ARCH with constant scaling (which is common in RoPE-based length extrapolation methods), and \ARCH with the attention logits scaling defined in Eq.~\ref{eq:logits-scaling}.}
    \label{fig:results-logits-scaling-from-scratch}
\end{figure}

Figure~\ref{fig:results-logits-scaling-from-scratch} report the results of \ARCH without attention logits scaling in HyPE, which is described in Eq.~\ref{eq:logits-scaling}, but is repeated here for convenience:
\begin{equation}
    \begin{aligned}
        \text{softmax}\left(\frac{s_t \mathbf q_t \mathbf K}{\sqrt{d_h}}\right) , \quad s_t &= \log_{a}(t + a),
    \end{aligned}
\end{equation}
As one can see from Figure~\ref{fig:results-logits-scaling-from-scratch}, without logits scaling~(i.e., setting $s_t=1$), HyPE exhibits limited length generalization abilities. With constant scaling~(setting $s_t=1.5$ for all positions) improve length generalization to a decent degree. But the full potential of HyPE is unlocked with a position-dependent scaling factor, setting $s_t = \log_a(t+a)$.

\subsubsection{\DISTILL Configuration Ablation Experiments}
\label{sec:conversion-ablation}

\begin{table*}[!h]
    \centering
    \footnotesize
    \caption{Ablation experiment results for stage 1 and 2 of \DISTILL, applied to Qwen3-1.7B.}
    \begin{tabular}{lc|cccccc}
        \toprule
        & & \multicolumn{6}{c}{\textbf{Needle-in-a-Haystack}} \\
        \cmidrule(lr){3-8}
        \textbf{Model} & \textbf{CSR} & \textbf{4K} & \textbf{8K} & \textbf{16K} & \textbf{32K} & \textbf{64K} & \textbf{128K}  \\
        \midrule
        \midrule
        \multicolumn{8}{c}{\textbf{Stage 1 ablations}} \\
        \midrule
        100M tokens (RADLADS) & 55.2 & 91.7 & 89.9 & 80.9 & 87.8 & 84.1 & 79.9 \\
        \rowcolor{orange!10}
        320M tokens (ours) & 55.4 & 95.5 & 93.3 & 88.2 & 92.5 & 87.3 & 80.9 \\
        625M tokens & 55.4 & 95.1 & 94.5 & 90.0 & 92.0 & 86.7 & 75.6 \\
        1.3B tokens & 55.1 & 90.5 & 89.5 & 81.0 & 91.4 & 83.1 & 61.2 \\
        % 2.5B tokens & 55.2 & 88.9 & 89.1 & 78.5 & 69.5 & 74.1 & 62.2 \\
        \midrule
        \midrule
        \multicolumn{8}{c}{\textbf{Stage 2 ablations}} \\
        \midrule
        % \multicolumn{8}{l}{\textit{20K train steps}}\\
        Max LR = 1e-5 (RADLADS)  & 46.9	& 89.2	& 72.7	& 71.2	& 88.1	& 65.1 &	60.7 \\
        Max LR = 3e-5         & 55.5	& 67.0	& 70.1	& 66.4	& 64.5	& 54.2 &	54.9 \\
        \rowcolor{orange!10} 
        Max LR = 1e-4 (ours)  & 56.4	& 79.9	& 75.4	& 76.9	& 78.7	& 70.1 &	68.7 \\
        Max LR = 3e-4         & 46.0	& 71.1	& 61.2	& 36.4	& 39.8	& 36.1 &	36.1 \\
        Max LR = 1e-3         & 36.8	& 79.2	& 73.9	& 75.5	& 75.1	& 84.5 &	75.1 \\
        % \midrule
        % \multicolumn{8}{l}{\textit{10K train steps, context length: 512 $\rightarrow$ 2048 (smaller batch size)}}\\
        % LR = 1e-4         & 55.9	& 79.2	& 73.9	& 75.5	& 75.1	& 67.0 &	65.5 \\
        % \midrule
        % \midrule
        % \multicolumn{8}{c}{\textbf{Stage 3 ablations}} \\
        % \midrule
        % ...            &  \\
        \bottomrule
    \end{tabular}
    \label{tab:training-ablation-results}
\end{table*}

Table~\ref{tab:training-ablation-results} presents the ablation experiments on the training configurations in our conversion procedure, \DISTILL. For stage 1, surprisingly, increasing the amount of training data beyond 320M tokens does not result in strong final performance. For stage 2, we can see that the default constant LR from RADLADS~\citep{radlads} is highly suboptimal. This discrepancy might be a result of the fact that RADLADS employs a different network architecture than ours and/or that their model sizes are different.

\section{Which RNN Mixers are Compatible with \ARCH?}
\label{sec:appendix-compatible-rnn-mixers}

Here, we describe a more comprehensive (but not exhaustive) list of RNN mixers that are compatible with \ARCH. In other words, they can be expressed as Eq.~(\ref{eq:update-rule}) and (\ref{eq:query-rule})), which we rewrite here for convenience.
\begin{equation}
    \begin{aligned}
    \mathbf q_t &= \mathbf x_t \mathbf W_q, \quad \mathbf k_t = \mathbf x_t \mathbf W_k,\quad \mathbf v_t = \mathbf x_t \mathbf W_v, \\
    \mathbf S_{t} &= \mathbf F_{t} \mathbf S_{t-1} + \mathbf k_t^\top \mathbf v_t \in \mathbb R^{d_h\times d_h}, \\
    \mathbf y_{t} &= \mathbf q_{t} \mathbf S_{t} \mathbf W_o^\top \in \mathbb R^{d},
    \end{aligned} \label{eq:outer-product-rnn}
\end{equation}
This formulation includes (but are not limited to) the RNN mixers listed in Table~\ref{tab:mixers-compatible-with-ours}. Table~\ref{tab:mixer-variable-mapping} describes how the notations from each of the mixers studied in this paper correspond to our notations for RNN mixers~(i.e., Eq.~(\ref{eq:outer-product-rnn})). It also illustrates which components in these RNN mixers inherit the attention weights in \DISTILL.

\begin{table}[!h]
    \centering
    \footnotesize
    \caption{Non-exhaustive list of representative RNN mixers that are compatible with \ARCH and \DISTILL.}
    \begin{tabular}{|l|l|}
        \hline
        Linear Attention~\citep{transformers-are-rnns}
            & RetNet~\citep{retnet} \\
        \hline
        Lightning Attention~\citep{lightning-attn}
            & HGRN-2~\citep{hgrn2} \\
        \hline
        GLA~\citep{gla}
            & Mamba2~\citep{mamba2} \\
        \hline
        GSA~\citep{gsa}
            & DeltaNet~\citep{deltanet} \\
        \hline
        GDN~\citep{gdn}
            & RWKV-7~\citep{rwkv7} \\
        \hline
        TTT~\citep{ttt}
            & Kimi DeltaAttention~\citep{kimi-linear} \\
        \hline
    \end{tabular}
    \label{tab:mixers-compatible-with-ours}
\end{table}

\begin{table}[!h]
    \centering
    \footnotesize
    \caption{List of how various state-of-the-art RNNs can be expressed as outer-product-based RNNs (i.e., Eq~(\ref{eq:outer-product-rnn})), using the notations from their respective original paper. ``-'' indicates that these variables are never described in their original papers, but they can be found in the implementations. Our code for converting each of these RNN mixers is publicly available.}
    \begin{tabular}{l|l|l|l|l|l|l|l|l}
        \toprule
        \rowcolor{gray!10}
        \textbf{Mixer}   & $\mathbf F_t$ & $\mathbf q_t$ & $\mathbf k_t$ & $\mathbf v_t$ & $\mathbf W_q$ & $\mathbf W_k$ & $\mathbf W_v$ & $\mathbf W_o$ \\
        \midrule
        Lightning Attention
            & $\lambda$ 
            & $\mathbf q_t$ 
            & $\mathbf k_t$ 
            & $\mathbf v_t$
            & $\mathbf W_q$
            & $\mathbf W_k$
            & $\mathbf W_v$
            & - \\
        \midrule
        Mamba2  
            & $\alpha_t I$ 
            & $C_t$ 
            & $\Delta_t B_t$
            & $x_t$
            & -
            & -
            & $W^{(x)}$
            & $W^{(o)}$\\
        \midrule
        GLA     
            & $\text{diag} \left(\alpha_t \right)$
            & $\bm{q}_t$
            & $\bm{k}_t$
            & $\bm{v}_t$
            & $\bm W_Q$
            & $\bm W_K$
            & $\bm W_V$
            & $\bm W_O$ \\
        \midrule
        GDN 
            & $\alpha_t \left(I - \beta_t \bm k_t^\top \bm k_t\right)$
            & $\bm q_t$
            & $\bm k_t$
            & $\bm v_t$ 
            & $\bm W_Q$
            & $\bm W_K$
            & $\bm W_V$
            & - \\
        \midrule
        RWKV-7  
            & $\left(\text{diag}(\omega_t) - \hat {\kappa}_t k_t^\top(a_t \odot \hat{\kappa}_t )\right)$ 
            & $r_t $
            & $\tilde k_t$
            & $\nu_t$ 
            & $\bm W_r$ 
            & $\bm W_k$
            & $\bm W_v$ 
            & $\bm W_o$\\
        \bottomrule
    \end{tabular}
    \label{tab:mixer-variable-mapping}
\end{table}

\subsection{\ARCH's Compatibility with Mamba2}
\label{sec:appendix-compatibility-with-mamba2}

Mamba2 is derived from the perspective of state space models (SSMs), which is not based on QKV as the input. State space models may not always be expressible as Eq~(\ref{eq:outer-product-rnn}). Fortunately, Mamba and Mamba2 are special cases of SSMs that can be expressed as gated linear attention~\citep{gla}. The Mamba2 paper~\citep{mamba2} provides an in-depth discussion of how (gated) linear attention is related to SSMs. In brief, both of these state-of-the-art SSMs are compatible with \ARCH.

\paragraph{Multi-Head Mechanism}
However, from the perspective of linear attention, Mamba2 adopts a multi-value mechanism in which all heads share the same set of queries and keys. This is not the usual configuration for softmax attention models. Therefore, in order to utilize the pre-trained model weights of softmax attention models, we use multi-head Mamba2 in this paper. This change has a negligible impact on the model's throughput. 

\subsection{A Note on Kimi Delta Attention}

Here, we discuss a failed attempt at converting Qwen3's attention into KDA~\citep{kimi-linear}, in order to facilitate more effctive research. We have tried to use \DISTILL to convert Qwen3's attention layers into KDA layers using the same configurations as described in Appendix~\ref{sec:appendix-training-configs}. However, the training process could not converge with the gradient norm becoming \texttt{inf} after a few steps in stage 2. We tried reducing the learning rate but it did not help.

\section{Training and Model Configurations for Training From Scratch Experiments}
\label{sec:appendix-training-from-scratch-configs}

Here, we describe the training and model configurations for the experiments in Section~\ref{sec:training-from-scratch-results}.

\begin{table}[!h]
    \centering
    \footnotesize
    \caption{Training configurations and hyperparameters used when training from scratch (Section~\ref{sec:training-from-scratch-results}).}
    \label{tab:training-configs}
    \begin{tabular}{lc}
        \toprule
        \textbf{Hyperparameter} & \textbf{Value} \\
        \midrule
        Total tokens    & 20B \\
        Context length  & 4096 \\
        Batch size      & 128 \\
        Training steps     & 40,000 \\
        LR scheduler    & WSD~\citep{wsd} \\
        Max. Learning rate & $5 \times 10^{-4}$ \\
        Min. learning rate & $5 \times 10^{-5}$\\
        LR warmup steps    & 1,000 \\
        LR decay steps     & 8,000 \\
        Optimizer       & AdamW, $\beta=(0.9, 0.95)$\\
        Weight decay    & 0.1 \\
        \bottomrule
    \end{tabular}
\end{table}

\begin{table}[!t]
    \footnotesize
    \centering
    \caption{Model architecture configurations for the from-scratch training experiments (Section~\ref{sec:training-from-scratch-results}). The tokenizer for all models is the GPT-2 tokenizer\footnote{\url{https://huggingface.co/openai-community/gpt2}}.}
    \label{tab:model-configs}
    \begin{tabular}{lccc}
        \toprule
        \textbf{Hyperparameter} & \textbf{Transformer} & \textbf{SWAN-GPT} & \textbf{\ARCH} \\
        \midrule
        Tokenizer       & GPT-2 & GPT-2 & GPT-2 \\
        Vocabulary size & 50,304 & 50,304 & 50,304 \\
        Layers          & 28    & 28    & 28 \\
        Hidden size     & 1024  & 1024  & 1024 \\
        RNN layers      & 0     & 0     & 21 \\
        Full Attn. layers & 28  & 7     & 7 \\
        SWA layers      & 0     & 21    & 0 \\
        SWA Window size & --    & 512   & -- \\
        FNN width       & 3072  & 3072  & 3072 \\
        Head dim        & 128   & 128   & 128 \\
        Attention heads & 16    & 16    & 16 \\
        Attention KV heads & 8  & 8     & 8 \\
        RNN heads       & --    & --    & 16 \\
        Tie embeddings  & Yes   & Yes   & Yes \\
        QK Norm in attention & Yes & Yes & Yes \\
        RoPE $\theta$   & 50k   & 50k   & 50k \\
        \bottomrule
    \end{tabular}
\end{table}

\subsection{Training Configurations}
\label{sec:appendix-training-from-scratch-training-configs}

All models are trained on 20 billion tokens from the FineWeb-edu dataset~\citep{fineweb-edu}. We use 8 NVIDIA A800 GPUs to train each model. The training code is based on the HuggingFace Accelerate framework. The specific training hyperparameters are detailed in Table~\ref{tab:training-configs}. The hyparameters are chosen to best match standard practicing in LLM pre-training.

\subsection{Model Configurations}
\label{sec:appendix-training-from-scratch-model-configs}

To ensure fair comparison, the parameter count for all models is controlled at approximately 500M. We also try to keep the implementation as similar as possible to its official implementation released by the respective authors. For \ARCH models, 25\% of the layers are attention layers, interleaved with RNN layers in a repeating pattern of one attention layer followed by three RNN layers (i.e., Attn $\to$ RNN $\to$ RNN $\to$ RNN)\footnote{Since we are training from scratch, we do not need to handle attention layer selection as in \DISTILL.}. The MLP blocks after each attention/RNN block are always a SwiGLU block with the same hyperparameters. Table~\ref{tab:model-configs} reports the detailed configuration for each model, Table~\ref{tab:scaling-hyperparameter-of-from-scratch-models} reports the attention logits scaling (see Section~\ref{sec:hype}) for each model., and Table~\ref{tab:from-scratch-rnn-hyperparams} reports the configurations for each RNN mixer. The ensure fair comparison with the Transformer model and SWAN-GPT and also to better compare with our \ARCH models that are distilled from pre-trained Transformer models, we do not employ short convolutions in RNN mixers.

\begin{table}[!t]
    \centering
    \footnotesize
    \caption{The logits scaling hyperparameter of various models in the from-scratch training experiments (Section~\ref{sec:training-from-scratch-results}).}
    \begin{tabular}{l|c}
        \toprule
        \textbf{Model}  &  \textbf{Logit scaling base $a$ (Eq.~\ref{eq:logits-scaling})} \\
        \midrule
        Transformer      & None \\
        \ARCH-Lightning & 300 \\
        \ARCH-Lightning (all NoPE) & 1000 \\
        \ARCH-GDN       & 200 \\
        \ARCH-GLA       & 500 \\
        \ARCH-RWKV7     & 5000 \\
        \ARCH-Mamba2    & 1000 \\
        SWAN-GPT        & 1000 \\
        \bottomrule
    \end{tabular}
    \label{tab:scaling-hyperparameter-of-from-scratch-models}
\end{table}

\clearpage
\makeatletter
\setlength{\@fptop}{0pt}
\makeatother

\begin{table}[!t]
    \setlength{\tabcolsep}{3pt}
    \centering
    \footnotesize
    \caption{Hyperparameters for of the RNN layers in the \ARCH variants of the from-scratch training experiments (Section~\ref{sec:training-from-scratch-results}). \ding{51} denotes that the feature is enabled, \ding{55} denotes disabled, and ``--'' means that the hyperparameter is not applicable.}
    \begin{tabular}{l|p{1.6cm}|p{1.8cm}|p{1.6cm}|p{1.6cm}|p{1.6cm}}
        \toprule
        \textbf{Hyperparameter} & \textbf{HypeNet-Lightning} & \textbf{HypeNet-GDN} & \textbf{\ARCH-GLA} & \textbf{HypeNet-RWKV7} & \textbf{HypeNet-Mamba2} \\
        \midrule
        \multicolumn{6}{l}{\textit{Gating \& Normalization}} \\
        \midrule
        Output gate         & \ding{51}     & \ding{51} & \ding{51} & \ding{51} & \ding{51}  \\
        Output norm         & \ding{51}     & \ding{51} & \ding{51} & \ding{51} & \ding{51}  \\
        QK norm             & \ding{51}     & \ding{51}, $L_2$-norm & \ding{55} & \ding{55} & \ding{55}  \\
        % \texttt{elementwise\_affine} & --   & --        & --        & \ding{51} & -- \\
        % \midrule
        % \multicolumn{6}{l}{\textit{Kernel \& Expansion Settings}} \\
        % \midrule
        % Head dim            & 128           & 128       & 128       & 128       & 128  \\
        QKV activation      & \ding{55}          & \ding{51}, SiLU      & \ding{55}      & \ding{55}      & \ding{55} \\
        Short Convolution   & \ding{55}     & \ding{55} & \ding{55} & \ding{55} & \ding{55} \\
        $\mathbf F_t$ neg. eigenvalue   
            & \ding{55}     & \ding{51}     & \ding{55} & \ding{55} & \ding{55} \\
        \midrule
        \multicolumn{6}{l}{\textit{Low-Rank Parametrization}} \\
        \midrule
        Gate low-rank dim.  & --            & --        & 16        & 160       & --  \\
        Value low-rank dim.      & --            & --        & --        & 96        & --  \\
        Decay low-rank dim.      & --            & --        & --        & 160       & --  \\
        $A$ low-rank dim.        & --            & --        & --        & 160       & -- \\
        % \midrule
        % Logit Scaling $a$ (Eq.~\ref{eq:logits-scaling}) & 300 & 200 & 500 & 5000 & 1000 & 1000 \\
        \bottomrule
    \end{tabular}%
    \label{tab:from-scratch-rnn-hyperparams}
\end{table}

\vspace*{\fill} % This creates flexible vertical space

% \section{You \emph{can} have an appendix here.}

% You can have as much text here as you want. The main body must be at most $8$
% pages long. For the final version, one more page can be added. If you want, you
% can use an appendix like this one.

% The $\mathtt{\backslash onecolumn}$ command above can be kept in place if you
% prefer a one-column appendix, or can be removed if you prefer a two-column
% appendix.  Apart from this possible change, the style (font size, spacing,
% margins, page numbering, etc.) should be kept the same as the main body.
%%%%%%%%%%%%%%%%%%%%%%%%%%%%%%%%%%%%%%%%%%%%%%%%%%%%%%%%%%%%%%%%%%%%%%%%%%%%%%%
%%%%%%%%%%%%%%%%%%%%%%%%%%%%%%%%%%%%%%%%%%%%%%%%%%%%%%%%%%%%%%%%%%%%%%%%%%%%%%%

\end{document}